\documentclass[runningheads]{llncs}
\usepackage{amsmath,graphicx}
\usepackage{bbold}
\usepackage{tabularray}
\usepackage{bm}
\usepackage{hyperref}
\usepackage{makecell}
\usepackage{float}
\usepackage{arydshln}
\usepackage{graphics}
\usepackage{multirow}
\usepackage{pifont}
\newcommand{\cmark}{\ding{51}}%
\newcommand{\xmark}{\ding{55}}%
\usepackage{amssymb}
\usepackage{subcaption}
\usepackage{soul}
\usepackage[dvipsnames]{xcolor}
\newcommand{\argmin}{\mathop{\mathrm{arg\,min}}}

\usepackage{url,wrapfig}
\usepackage{lipsum} 
\newcommand{\uls}[1]{\underline{\smash{}}}
\begin{document}

\title{Weakly-supervised positional contrastive learning: application to cirrhosis classification}

\titlerunning{Weakly-supervised positional contrastive learning}

\author{Emma Sarfati\inst{1,2}
\and
Alexandre Bône\inst{1}
\and
Marc-Michel Rohé\inst{1}
\and 
Pietro Gori\inst{2}
\and 
Isabelle Bloch\inst{2,3}
}

\authorrunning{E. Sarfati et al.}
%
\institute{Guerbet Research, Villepinte, France \and
LTCI, Télécom Paris, Institut Polytechnique de Paris, France\and
Sorbonne Université, CNRS, LIP6, Paris, France}
%
\maketitle              
\begin{abstract} 
Large medical imaging datasets can be cheaply and quickly annotated with low-confidence, weak labels (\textit{e.g.}, radiological scores). Access to high-confidence labels, such as histology-based diagnoses, is rare and costly. Pretraining strategies, 
like contrastive learning (CL) methods, can leverage unlabeled or weakly-annotated datasets. These methods typically require large batch sizes, which poses a difficulty in the case of large 3D images at full resolution, due to limited GPU memory. Nevertheless, volumetric positional information about the spatial context of each 2D slice can be very important for some medical applications. In this work, we propose an efficient weakly-supervised positional (WSP) contrastive learning strategy where we integrate both the spatial context of each 2D slice and a weak label via a generic kernel-based loss function. We illustrate our method on cirrhosis prediction using a large volume of weakly-labeled images, namely radiological low-confidence annotations, and small strongly-labeled (\textit{i.e.}, high-confidence) datasets. The proposed model improves the classification AUC by 5\% with respect to a baseline model on our internal dataset, and by 26\% on the public LIHC dataset from the Cancer Genome Atlas. The code is available at: \url{https://github.com/Guerbet-AI/wsp-contrastive}.
\end{abstract}

\section{Introduction}

In the medical domain, obtaining a large amount of high-confidence labels, such as histopathological diagnoses, is arduous due to the cost and required technicality. It is however possible to obtain lower confidence assessments for a large amount of images, either by a clinical questioning, or directly by a radiological diagnosis. To take advantage of large volumes of unlabeled or weakly-labeled images, pre-training encoders with self-supervised methods showed promising results in deep learning for medical imaging \cite{multiinstance,localcontrastive,clmedical,sslmedical,modelgenesis,rubikcube}. In particular, contrastive learning (CL) is a self-supervised method that learns a mapping of the input images to a representation space where similar (positive) samples are moved closer and different (negative) samples are pushed far apart.
Weak discrete labels can be integrated into contrastive learning by, for instance, considering as positives only the samples having the same label, as in~\cite{supcon}, or by directly weighting unsupervised contrastive and supervised cross entropy loss functions, as in~\cite{sarfati}.
In this work, we focus on the scenario where radiological meta-data (thus, low-confidence labels) are available for a large amount of images, whereas high-confidence labels, obtained by histological analysis, are scarce.

Naive extensions of contrastive learning methods, such as~\cite{simclr,byol,moco}, from 2D to 3D images may be difficult due to limited GPU memory and therefore small batch size. A usual solution consists in using patch-based methods~\cite{yaware,patch}. However, these methods pose two difficulties: they reduce the spatial context (limited by the size of the patch), and they require similar spatial resolution across images. This is rarely the case for abdominal CT/MRI acquisitions, which are typically strongly anisotropic and with variable resolutions. Alternatively, depth position of each 2D slice, within its corresponding volume, can be integrated in the analysis. For instance, in~\cite{localcontrastive}, the authors proposed to integrate depth in the sampling strategy for the batch creation. Likewise, in~\cite{positional}, the authors proposed to define as similar only 2D slices that have a \textit{small} depth difference, using a normalized depth coordinate $d\in[0,1]$. These works implicitly assume a certain threshold on depth to define positive and negative samples, which may be difficult to define and may be different among applications and datasets. Differently, inspired by \cite{carlo,yaware}, here we propose to use a degree of “positiveness”
between samples by defining a kernel function $w$ on depth positions. This allows us to consider volumetric depth information during pre-training \textit{and} to use large batch sizes. Furthermore, we also propose to \textit{simultaneously} leverage weak discrete attributes during pre-training by using a novel and efficient contrastive learning composite kernel loss function, denoting our global method Weakly-Supervised Positional (WSP).

We apply our method to the classification of histology-proven liver cirrhosis, with a large volume of (weakly) radiologically-annotated CT-scans and a small amount of histopathologically-confirmed cirrhosis diagnosis. We compare the proposed approach to existing self-supervised methods.

\section{Method}
Let $x_t$ be an input 2D image, usually called \textit{anchor}, extracted from a 3D volume, $y_t$ a corresponding discrete weak variable and $d_t$ a related continuous variable. In this paper, $y_t$ refers to a weak radiological annotation and $d_t$ corresponds to the normalized depth position of the 2D image within its corresponding 3D volume: if $V_{max}$ corresponds to the 
maximal depth-coordinate of a volume $V$, we compute $d_t=\frac{p_t}{V_{max}}$ with $p_t\in[0,V_{max}]$ being the original depth coordinate. Let $x_j^-$ and $x_i^+$ be two semantically different (negative) and similar (positive) images with respect to $x_t$, respectively. 

The definition of similarity is crucial in CL and is the main difference between existing methods.
For instance, in unsupervised CL, methods such as SimCLR~\cite{simclr,simclr2} choose as positive samples random augmentations of the anchor $x_i^+=t(x_t)$, where $t\sim \mathcal{T}$ is a random transformation chosen among a user-selected family~$\mathcal{T}$. Negative images $x_j^-$ are all other (transformed) images present in the batch.  

Once $x_j^-$ and $x_i^+$ are defined, the goal of CL is to compute a mapping function $f_\theta: \mathcal{X} \rightarrow \mathbb{S}^d$, where $\mathcal{X}$ is the set of images and $\mathbb{S}^d$ the representation space, so that similar samples are mapped closer in the representation space than dissimilar samples. Mathematically, this can be defined as looking for a $f_\theta$ that satisfies the condition: 
\begin{equation}
s_{tj}^- - s_{ti}^+ \leq 0 \quad \forall t,j,i 
\label{clglobal}
\end{equation}
where $s_{tj}^-=sim(f_\theta(x_t),f_\theta(x_j^-))
$ and $s_{ti}^+=sim(f_\theta(x_t),f_\theta(x_i^+))
$, with $sim$ a similarity function defined here as $sim(a,b)=\frac{a^Tb}{\tau}$ with $\tau>0$.

In the presence of discrete labels $y$, the definition of negative ($x_j^-$) and positive ($x_i^+$) samples may change. For instance, in SupCon~\cite{supcon}, the authors define as positives all images with the same discrete label $y$. However, when working with continuous labels $d$, one cannot use the same strategy since all images are somehow positive and negative at the same time. A possible solution \cite{positional} would be to define a threshold $\gamma$ on the distance between labels (\textit{e.g.}, $d_a$, $d_b$) so that, if the distance is smaller than $\gamma$ (\textit{i.e.}, $||d_a - d_b||_2 < \gamma$), the samples (\textit{e.g.}, $x_a$ and $x_b$) are considered as positives. However, this requires a user-defined hyper-parameter $\gamma$, which could be hard to find in practice. A more efficient solution, as proposed in~\cite{yaware}, is to define a degree of ``positiveness'' between samples using a normalized kernel function $w_{\sigma}(d,d_i) = K_{\sigma}(d - d_i)$, where $K_{\sigma}$ is, for instance, a Gaussian kernel, with user defined hyper-parameter $\sigma$ and $0 \leq w_{\sigma} \leq 1$. It is interesting to notice that, for discrete labels, one could also define a kernel as:  $w_{\delta}(y,y_i) = \delta(y - y_i)$, $\delta$ being the Dirac function, retrieving exactly SupCon \cite{supcon}.

\begin{figure*}[t]
    \centering
    \includegraphics[width=11cm]{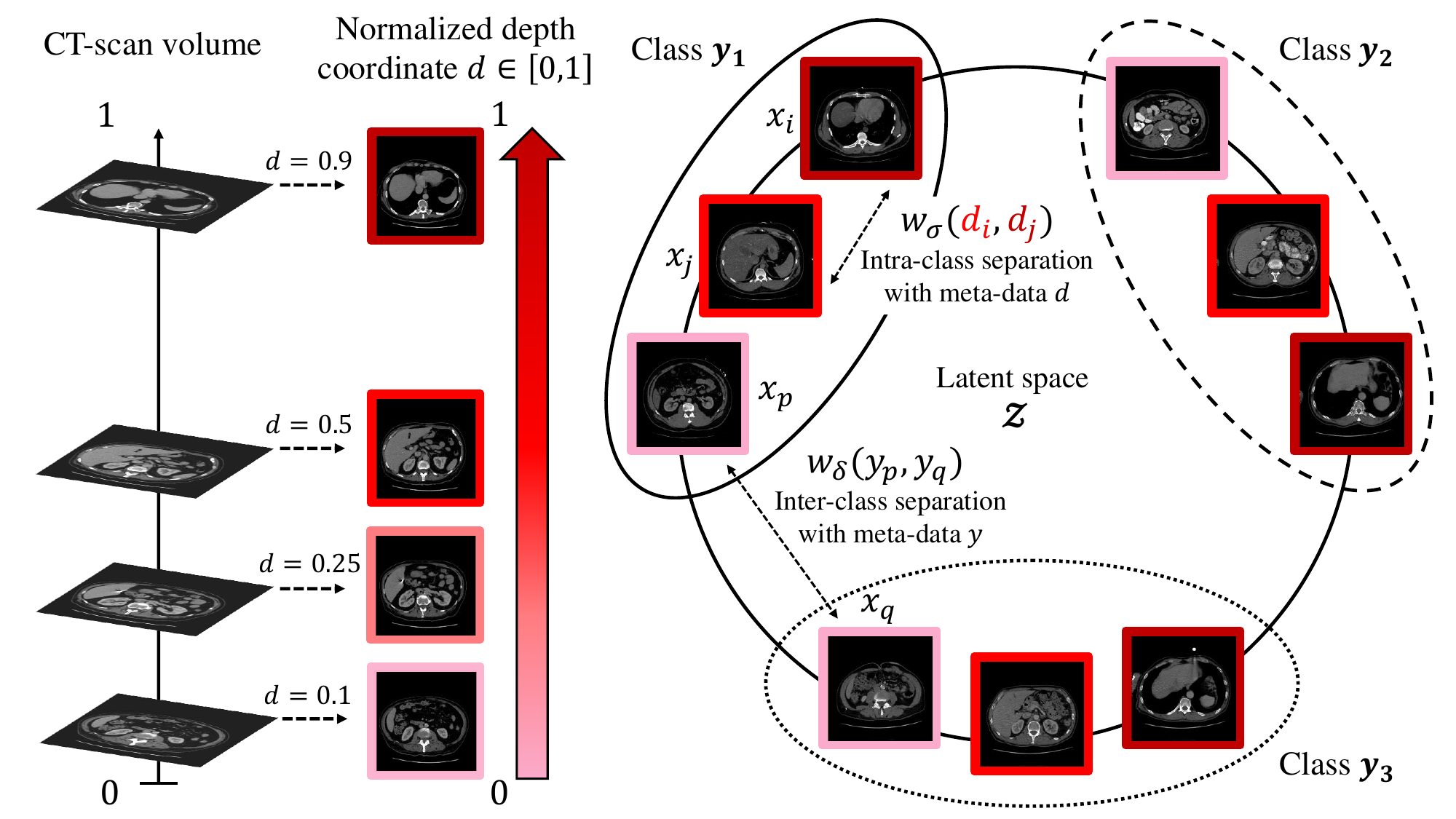}
    \caption{Example of representation space constructed by our loss function, leveraging both continuous depth coordinate $d$ and discrete label $y$ (\textit{i.e.}, radiological diagnosis $y_{radio}$). Samples from  different radiological classes are well separated and, at the same time, samples are ordered within each class based on their depth coordinate $d$.}
    \label{global}
\end{figure*}

In this work, we propose to leverage both continuous $d$ and discrete $y$ labels, by combining (here by multiplying) the previously defined kernels, $w_\sigma$ and $w_\delta$, into a composite kernel loss function. In this way, samples will be considered as similar (positive) only if they have a \textit{composite} degree of ``positiveness'' greater than zero, namely both kernels have a value greater (or different) than 0 ($w_\sigma > 0$ and $w_\delta \neq 0$). An example of resulting representation space is shown in Figure~\ref{global}. This constraint can be defined by slightly modifying the condition introduced in Equation~\ref{clglobal}, as:
\begin{equation}
\underbrace{w_\delta(y_t,y_i) \cdot w_\sigma(d_t,d_i)}_{\text{composite kernel } w_{ti}} (s_{tj}-s_{ti}) \leq 0 \quad \forall t,i,j\neq i
\label{ours}
\end{equation}
where the indices $t,i,j$ traverse all $N$ images in the batch since there are no ``hard'' positive or negative samples, as in SimCLR or SupCon, but all images are considered as positive and negative at the same time. As commonly done in CL~\cite{barbano_unbiased_2023}, this condition can be transformed into an optimization problem using the $\max$ operator and its smooth approximation \textit{LogSumExp}:
\begin{equation}
    \begin{gathered}
        \argmin_{f_\theta} \sum_{t,i} \max(0, w_{ti} \{s_{tj} - s_{ti} \}_{\substack{j=1 \\ j \neq i}}^N) = \argmin_{f_\theta}\sum_{t,i} w_{ti} \max(0, \{ s_{tj} - s_{ti} \}_{\substack{j=1 \\ j \neq i}}^N) \\ 
        \approx \argmin_{f_\theta} \left( - \sum_{t,i} w_{ti} \log \left( \frac{\exp(s_{ti})}{\sum_{j \neq i}^N \exp(s_{tj})}  \right)\right)
    \end{gathered}
\end{equation}

By defining $P(t)=\{i:y_i=y_t\}$ as the set of indices of images $x_i$ in the batch with the same discrete label $y_i$ as the anchor $x_t$, we can rewrite our final loss function as:
\begin{equation}
\mathcal{L}_{WSP}=-\sum_{t=1}^{N} \sum_{i\in P(t)} w_\sigma(d_t,d_i) \log \left( \frac{\exp(s_{ti})}{ \sum_{j\neq i}^{N} \exp(s_{tj})} \right)
\end{equation} 
where $w_\sigma(d_t,d_i)$ is normalized over $i\in P(t)$. In practice, it is rather easy to find a good value of $\sigma$, as the proposed kernel method is quite robust to its variation. A robustness study is available in the supplementary material. For the experiments, we fix $\sigma=0.1$.

\section{Experiments}
We compare the proposed method with different contrastive and non-contrastive methods, that either use no meta-data (SimCLR \cite{simclr}, BYOL \cite{byol}), or leverage only discrete labels (SupCon \cite{supcon}), or continuous labels (depth-Aware \cite{yaware}). The proposed method is the only one that takes simultaneously into account both discrete and continuous labels. In all experiments, we work with 2D slices rather than 3D volumes due to the anisotropy of abdominal CT-scans in the depth direction and the limited spatial context or resolution obtained with 3D patch-based or downsampling methods, respectively, which strongly impacts the cirrhosis diagnosis that is notably based on the contours irregularity. Moreover, the large batch sizes necessary in contrastive learning can not be handled in 3D due to a limited GPU memory.

\subsection{Datasets} 
Three datasets of abdominal CT images are used in this study. One dataset is used for contrastive pretraining, and the other two for evaluation. All images have a 512x512 size, and we clip the intensity values between -100 and 400. 
\\\noindent
$\bm{\mathcal{D}_{radio}}$.\quad First, $\mathcal{D}_{radio}$ contains 2,799 CT-scans of patients in portal venous phase with a radiological (weak) annotation, \textit{i.e.} realized by a radiologist, indicating four different stages of cirrhosis: no cirrhosis, mild cirrhosis, moderate cirrhosis and severe cirrhosis ($y_{radio}$). The respective numbers are 1880, 385, 415 and 119. $y_{radio}$ is used as the discrete label $y$ during pre-training.
\\\noindent
$\bm{\mathcal{D}_{histo}^1}$.\quad It contains 106 CT-scans from different patients in portal venous phase, with an identified histopathological status (METAVIR score) obtained by a histological analysis, designated as $y_{histo}^1$. It corresponds to absent fibrosis (F0), mild fibrosis (F1), significant fibrosis (F2), severe fibrosis (F3) and cirrhosis (F4). This score is then binarized to indicate the absence or presence of advanced fibrosis~\cite{li}: F0/F1/F2 (N=28) vs. F3/F4 (N=78).
\\\noindent
$\bm{\mathcal{D}_{histo}^2}$.\quad This is the public LIHC dataset from the Cancer Genome Atlas~\cite{lihc}, which presents a histological score, the Ishak score, designated as $y_{histo}^2$, that differs from the METAVIR score present in $\mathcal{D}_{histo}^1$. This score is also distributed through five labels: No Fibrosis, Portal Fibrosis, Fibrous Speta, Nodular Formation and Incomplete Cirrhosis and Established Cirrhosis. Similarly to the METAVIR score in $\mathcal{D}_{histo}^1$, we also binarize the Ishak score, as proposed in~\cite{metavirishak2,metavirishak}, which results in two cohorts of 34 healthy and 15 pathological patients.

In all datasets, we select the slices based on the liver segmentation of the patients. To gain in precision, we keep the top 70\% most central slices with respect to liver segmentation maps obtained manually in $\mathcal{D}_{radio}$, and automatically for $\mathcal{D}_{histo}^1$ and $\mathcal{D}_{histo}^2$ using a U-Net architecture pretrained on $\mathcal{D}_{radio}$ \cite{unet}. For the latter pretraining dataset, it presents an average slice spacing of 3.23mm with a standard deviation of 1.29mm. For the $x$ and $y$ axis, the dimension is
0.79mm per voxel on average, with a standard deviation of 0.10mm.

\subsection{Architecture and optimization.} 
\textbf{Backbones.} \quad We propose to work with two different backbones in this paper: TinyNet and ResNet-18~\cite{resnet}. TinyNet is a small encoder with 1.1M parameters, inspired by~\cite{yin}, with five convolutional layers, a representation space (for downstream tasks) of size 256 and a latent space (after a projection head of two dense layers) of size 64. In comparison, ResNet-18 has 11.2M parameters, a representation space of dimension 512 and a latent space of dimension 128. More details and an illustration of TinyNet are available in the supplementary material, as well as a full illustration of the algorithm flow.

\noindent \textbf{Data augmentation, sampling and optimization.} \quad CL methods~\cite{simclr,byol,moco} require strong data augmentations on input images, in order to strengthen the association between positive samples~\cite{augmentations}. In our work, we leverage three types of augmentations: rotations, crops and flips. Data augmentations are computed on the GPU, using the Kornia library~\cite{kornia}. During inference, we remove the augmentation module to only keep the original input images.
\\\indent
For sampling, inspired by~\cite{localcontrastive}, we propose a strategy well-adapted for contrastive learning in 2D medical imaging. We first sample $N$ patients, where $N$ is the batch size, in a balanced way with respect to the radiological/histological classes; namely, we roughly have the same number of subjects per class. Then, we randomly select only one slice per subject. In this way, we maximize the slice heterogeneity within each batch. We use the same sampling strategy also for classification baselines. For $\mathcal{D}_{histo}^2$, which has fewer patients than the batch size, we use a balanced sampling strategy with respect to the radiological/histological classes with no obligation of one slice per patient in the batch. As we work with 2D slices rather than 3D volumes, we compute the average probability per patient of having the pathology. The evaluation results presented later are based on the patient-level aggregated prediction. 
\\\indent
Finally, we run our experiments on a Tesla V100 with 16GB of RAM and a 6 CPU cores, and we used the PyTorch-Lightning library 
to implement our models. All models share the same data augmentation module, with a batch size of $B=64$ and a fixed number of epochs $n_{epochs}=200$. For all experiments, we fix a learning rate (LR) of $\alpha=10^{-4}$  and a weight decay of $\lambda=10^{-4}$. We add a cosine decay learning rate scheduler~\cite{cosine} to prevent over-fitting. For BYOL, we initialize the moving average decay at 0.996.

\noindent \textbf{Evaluation protocol.}\quad 
We first pretrain the backbone networks on $\mathcal{D}_{radio}$ using all previously listed contrastive and non-contrastive methods. Then, we train a regularized logistic regression on the frozen representations of the datasets $\mathcal{D}_{histo}^1$ and $\mathcal{D}_{histo}^2$. We use a stratified 5-fold cross-validation.
As a baseline, we train a classification algorithm from scratch (supervised) for each dataset, $\mathcal{D}_{histo}^1$ and $\mathcal{D}_{histo}^2$, using both backbone encoders and the same 5-fold cross-validation strategy. We also train a regularized logistic regression on representations obtained with a random initialization as a second baseline (random). Finally, we report the cross-validated results for each model on the aggregated dataset $\mathcal{D}_{histo}^{1+2}=\mathcal{D}_{histo}^1+\mathcal{D}_{histo}^2$.

\section{Results and Discussion} \label{sec:results}
\begin{table}[ht]
\caption{Resulting 5-fold cross-validation AUCs. For each encoder, best results are in \textbf{bold}, second top results are \underline{underlined}. * = We use the pretrained weights from ImageNet with ResNet-18 and run a logistic regression on the frozen representations. \\}\label{results}
\resizebox{\columnwidth}{!}{%
\begin{tabular}{c|c|c|c|c|c|c}
\textbf{Backbone}
 &
  \textbf{\begin{tabular}[c]{@{}c@{}}Pretraining \\ method\end{tabular}} &
  \multicolumn{1}{c|}{\begin{tabular}[c]{@{}c@{}}\textbf{Weak} \\ \textbf{labels}\end{tabular}} &
  \multicolumn{1}{c|}{\begin{tabular}[c]{@{}c@{}}\textbf{Depth}\\ \textbf{pos.}\end{tabular}} &
  $\bm{\mathcal{D}_{histo}^1}$ (N=106) &
  $\bm{\mathcal{D}_{histo}^2}$ (N=49) &
  $\bm{\mathcal{D}_{histo}^{1+2}}$ (N=155) \\ \hline
\multirow{6}{*}{\begin{tabular}[c]{@{}c@{}}TinyNet\end{tabular}} & Supervised        & \xmark & \xmark & 0.79 ($\pm$0.05) & 0.65 ($\pm$0.25) &      0.71 ($\pm$0.04)     \\ 
\hdashline
& None (random)      & \xmark & \xmark & 0.64 ($\pm$0.10) & 0.75 ($\pm$0.13) &  0.73 ($\pm$0.06)           \\ 
                  & SimCLR      & \xmark & \xmark & 0.75 ($\pm$0.08) & 0.88 ($\pm$0.16) & 0.76 ($\pm$0.11) \\ 
                  & BYOL        & \xmark & \xmark & 0.75 ($\pm$0.09) & \textbf{0.95 ($\pm$0.07)} & \underline{\smash{0.77 ($\pm$0.08)}}          \\ 
                  & SupCon      & \cmark & \xmark & 0.76 ($\pm$0.09) & \underline{\smash{0.93 ($\pm$0.07)}} & 0.72 ($\pm$0.06)          \\ 
                  & depth-Aware & \xmark & \cmark & \underline{\smash{0.80 ($\pm$0.13)}} & 0.81 ($\pm$0.08) & \underline{\smash{0.77 ($\pm$0.08)}}          \\ 
                  & Ours        & \cmark & \cmark & \textbf{0.84 ($\pm$0.12)} & 0.91 ($\pm$0.11) & \textbf{0.79 ($\pm$0.11)}\\ \Xhline{2\arrayrulewidth}
\multirow{7}{*}{\begin{tabular}[c]{@{}c@{}}ResNet-18\end{tabular}} & Supervised        &  \xmark & \xmark & 0.77 ($\pm$0.10) & 0.56 ($\pm$0.29) &   0.72 ($\pm$0.08)   \\ 
\hdashline
 & None (random)      & \xmark & \xmark & 0.69 ($\pm$0.19) & 0.73 ($\pm$0.12) &  0.68 ($\pm$0.09)           \\ 
 & ImageNet*      & \xmark & \xmark & 0.72 ($\pm$0.17) & 0.76 ($\pm$0.04) &  0.66 ($\pm$0.10)           \\ 
                   & SimCLR      & \xmark & \xmark & 0.79 ($\pm$0.09) & \underline{\smash{0.82 ($\pm$0.14)}} & 0.79 ($\pm$0.08)           \\ 
                   & BYOL        & \xmark & \xmark &           0.78 ($\pm$0.09) & 0.77 ($\pm$0.11) & 0.78 ($\pm$0.08)           \\ 
                   & SupCon      & \cmark & \xmark & 0.69 ($\pm$0.07) & 0.69 ($\pm$0.13) & 0.76 ($\pm$0.12) \\ 
                   & depth-Aware & \xmark & \cmark & \underline{\smash{0.83 ($\pm$0.07)}} & \underline{\smash{0.82 ($\pm$0.11)}} & \underline{\smash{0.80 ($\pm$0.07)}}  \\ 
                   & Ours        & \cmark & \cmark & \textbf{0.84 ($\pm$0.07)} & \textbf{0.85 ($\pm$0.10)} & \textbf{0.84 ($\pm$0.07)}\\ 
\end{tabular}%
}
\end{table}
\begin{figure}[!ht]
    \centering
    \includegraphics[width=11cm]{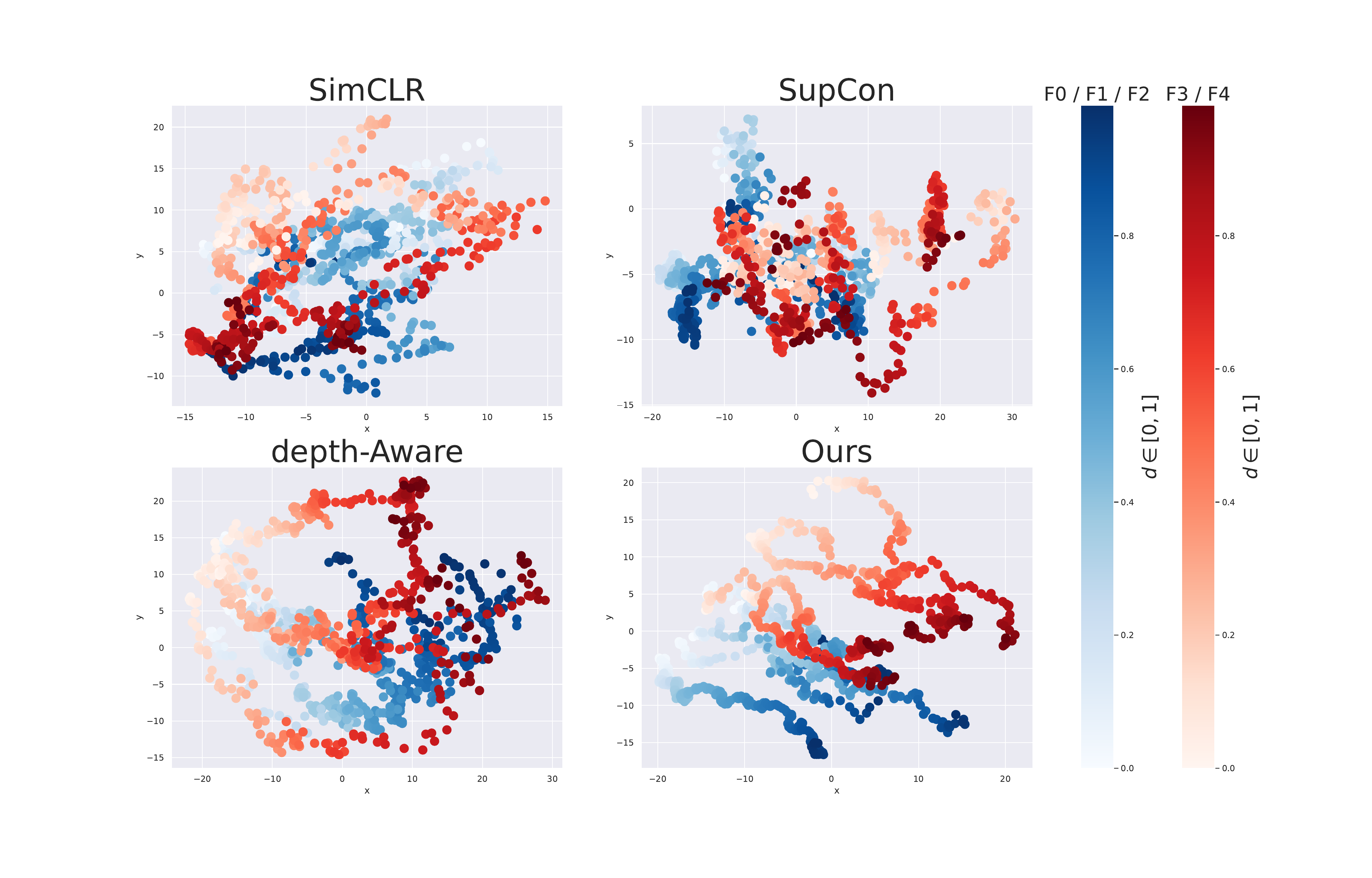}
    \caption{Projections of the ResNet-18 representation vectors of 10 randomly selected subjects of $\mathcal{D}_{histo}^1$ onto the first two modes of a PCA. Each dot represents a 2D slice. Color gradient refers to different depth positions. Red = cirrhotic cases. Blue = healthy subjects.
    }
    \label{pca}
\end{figure}
\noindent We present in Table~\ref{results} the results of all our experiments. For each of them, we report whether the pretraining method integrates the weak label meta-data, the depth spatial encoding, or both, which is the core of our method. First, we can notice that our method outperforms all other pretraining methods in $\mathcal{D}_{histo}^1$ and $\mathcal{D}_{histo}^{1+2}$, which are the two datasets with more patients. For the latter, the proposed method surpasses the second best pretraining method, depth-Aware, by 4\%. For $\mathcal{D}_{histo}^1$, it can be noticed that WSP (ours) provides the best AUC score whatever the backbone used. For the second dataset $\mathcal{D}_{histo}^2$, our method is on par with BYOL and SupCon when using a small encoder and outperforms the other methods when using a larger backbone. 

To illustrate the impact of the proposed method, we report in Figure~\ref{pca} the projections of the ResNet-18 representation vectors of 10 randomly selected subjects of $\mathcal{D}_{histo}^1$ onto the first two modes of a PCA. It can be noticed that the representation space of our method is the only one where the diagnostic label (not available during pretraining) and the depth position are correctly integrated. Indeed, there is a clear separation between slices of different classes (healthy at the bottom and cirrhotic cases at the top) and at the same time it seems that the depth position has been encoded in the $x$-axis, from left to right. SupCon performs well on the training set of $\mathcal{D}_{radio}$ (figure available in the supplementary material), as well as $\mathcal{D}_{histo}^2$ with TinyNet, but it poorly generalizes to $\mathcal{D}_{histo}^1$ and $\mathcal{D}_{histo}^{1+2}$. The method depth-Aware manages to correctly encode the depth position but not the diagnostic class label.

To assess the clinical performance of the pretraining methods, we also compute the balanced accuracy scores (bACC) of the trained classifiers, which is compared in Table~\ref{tab:bacc-results} to the bACC achieved by radiologists who were asked to visually assess the presence or absence of cirrhosis for the N=106 cases of $\mathcal{D}_{histo}^1$.

\begin{wraptable}{l}{7cm}
\vspace{-4mm}
\caption{Comparison of the pretraining methods with a binary radiological annotation for cirrhosis on $\mathcal{D}_{histo}^1$. Best results are in \textbf{bold}, second top results are \underline{underlined}.}
\label{tab:bacc-results}
\begin{tabular}{c|c|c}
\textbf{\begin{tabular}[c]{@{}c@{}}Pretraining \\ method\end{tabular}} & \textbf{\begin{tabular}[c]{@{}c@{}}bACC \\ models\end{tabular}} & \textbf{\begin{tabular}[c]{@{}c@{}}bACC \\ radiologists\end{tabular}} \\ \hline
Supervised        &      0.78 ($\pm$0.04)     & \multirow{7}{*}{} \\ 
\hdashline
None (random)      &      0.71 ($\pm$0.13)     &                   \\ 
ImageNet      &      0.74 ($\pm$0.13)     &                   \\ 
SimCLR      &      0.78 ($\pm$0.08)     &                   \\ 
BYOL        &      0.77 ($\pm$0.04)      &        0.82              \\ 
SupCon      &      0.77 ($\pm$0.10)     &                \\ 
depth-Aware &      \underline{\smash{0.84 ($\pm$0.04)}}     &                   \\ 
Ours        &      \textbf{0.85 ($\pm$0.09)} &                   \\ 
\end{tabular}
\vspace{-4mm}
\end{wraptable}

\noindent

\noindent The reported bACC values correspond to the best scores among those obtained with Tiny and ResNet encoders. Radiologists achieved a bACC of 82\% with respect to the histological reference. The two best-performing methods surpassed this score: depth-Aware and the proposed WSP approach, improving respectively the radiologists score by 2\% and 3\%, suggesting that including 3D information (depth) at the pretraining phase was beneficial. \\

\section{Conclusion}

In this work, we proposed a novel kernel-based contrastive learning method that leverages both continuous and discrete meta-data for pretraining. We tested it on a challenging clinical application, cirrhosis prediction, using three different datasets, including the LIHC public dataset. To the best of our knowledge, this is the first time that a pretraining strategy combining different kinds of meta-data has been proposed for such application. Our results were compared to other state-of-the-art CL methods well-adapted for cirrhosis prediction. The pretraining methods were also compared visually, using a 2D projection of the representation vectors onto the first two PCA modes. Results showed that our method has an organization in the representation space that is in line with the proposed theory, which may explain its higher performances in the experiments. As future work, it would be interesting to adapt our kernel method to non-contrastive methods, such as SimSIAM \cite{simsiam}, BYOL \cite{byol} or Barlow Twins \cite{barlowtwins}, that need smaller batch sizes and have shown greater perfomances in computer vision tasks.
In terms of application, our method could be easily translated to other medical problems, such as pancreas cancer prediction using the presence of intrapancreatic fat, diabetes mellitus or obesity as discrete meta-labels.

\noindent\textbf{Compliance with ethical standards.} \quad This research study was conducted retrospectively using human data collected from various medical centers, whose Ethics Committees granted their approval. Data was de-identified and processed according to all applicable privacy laws and the Declaration of Helsinki.

\noindent\textbf{Acknowledgments.} \quad This work was supported by Région Ile-de-France (ChoTherIA project) and ANRT (CIFRE \#2021/1735).

\bibliographystyle{splncs04}
\bibliography{bib_miccai}

\newpage
\appendix
\section{Supplementary Material}

\begin{figure}[!ht]
    \centering \includegraphics[width=12cm]{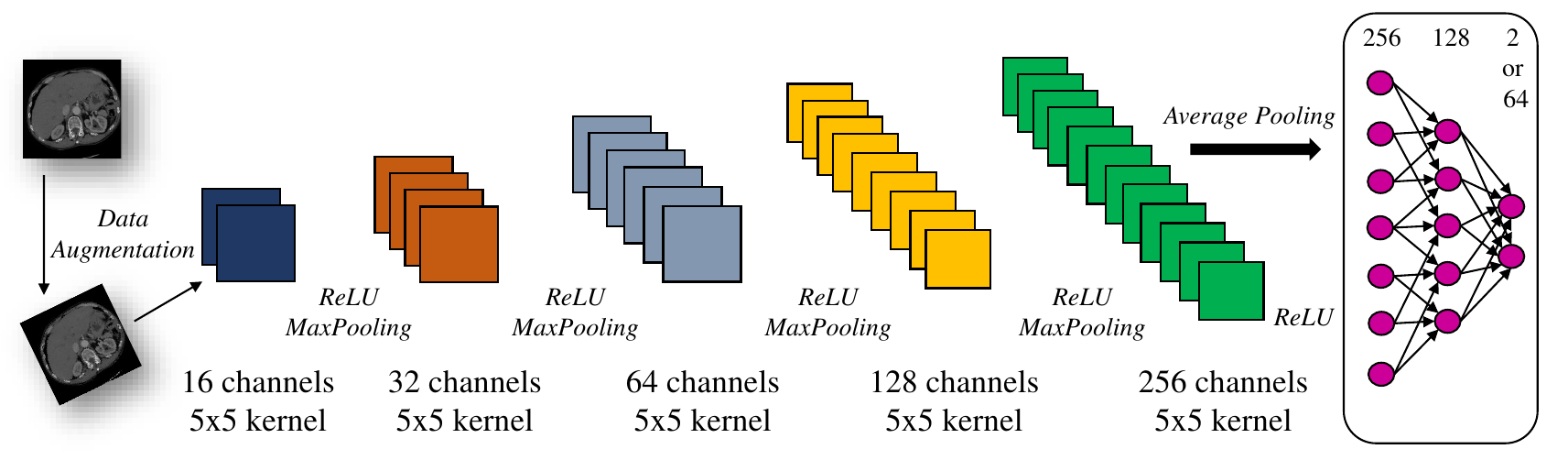}
    \caption{The proposed TinyNet used in our experiments.}
    \label{net}
\end{figure}

\begin{figure}[!ht]
    \centering \includegraphics[width=12cm]{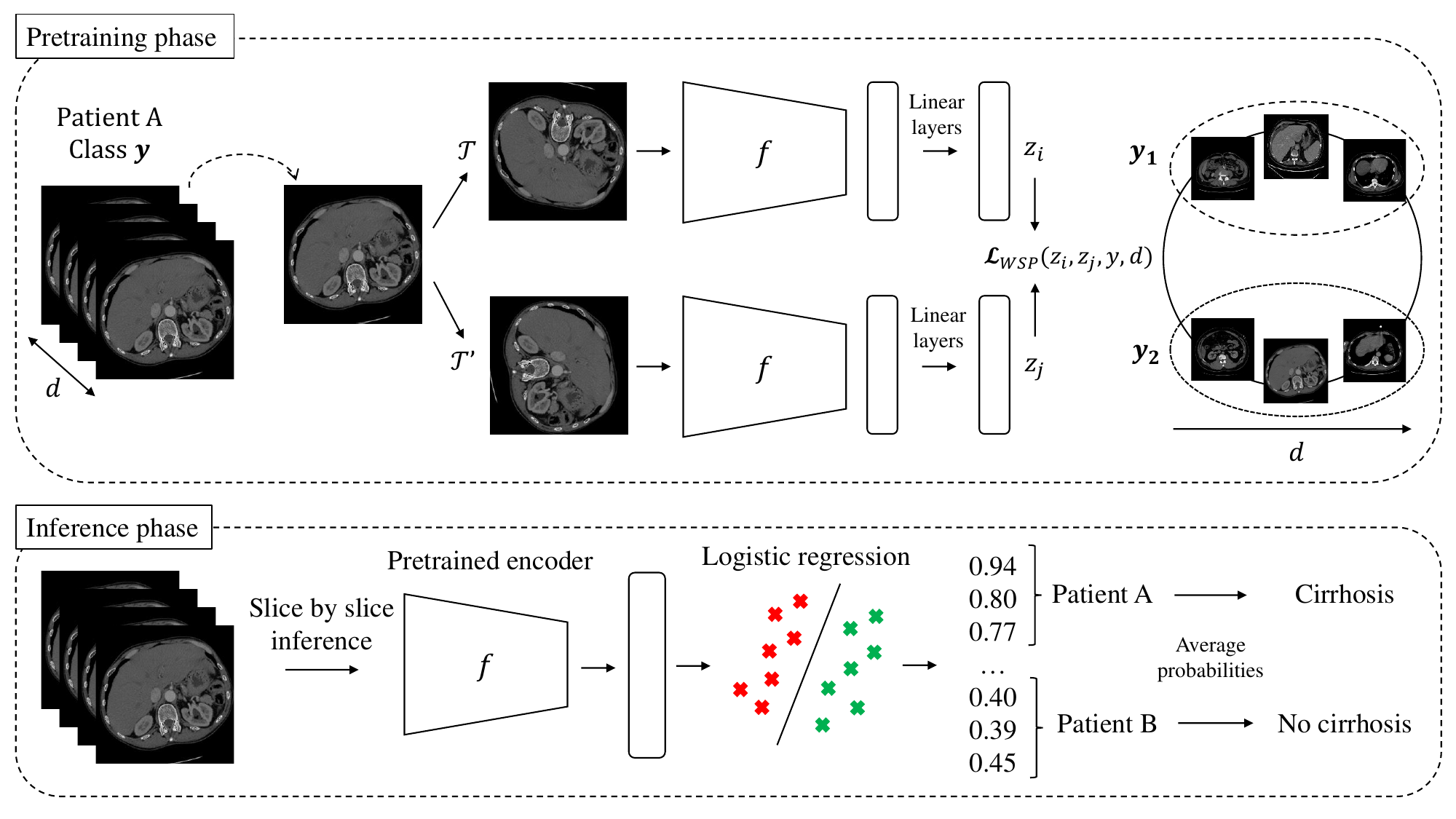}
    \caption{The full workflow of our method.}
    \label{net2}
\end{figure}


\begin{table}[!ht]
\caption{Resulting 5-fold cross-validation AUCs of the proposed method using the TinyNet backbone, varying the value of $\sigma$. In the paper, we chose the value of $\sigma=0.1$. One can interpret $\sigma$ as the proportion of slices around the anchor with a high weight accordance. The higher the value of $\sigma$ is, the more slices will be assigned a high weight value. \\}\label{robust}
\centering
\begin{tabular}{c|c|c|c}{\begin{tabular}[c]{@{}c@{}}$\bm{\sigma}$\end{tabular}} &
  $\bm{\mathcal{D}_{histo}^1}$ (N=106) &
  $\bm{\mathcal{D}_{histo}^2}$ (N=49) & $\bm{\mathcal{D}_{histo}^{1+2}}$ (N=155) \\ \hline
0.01        &  \multicolumn{1}{c|}{\underline{\smash{0.81 ($\pm$0.07)}}} & \multicolumn{1}{c|}{0.85 ($\pm$0.13)} & \textbf{0.81 ($\pm$0.08)}\\ 
0.1      &  \multicolumn{1}{c|}{\textbf{0.85 ($\pm$0.10)}} & \multicolumn{1}{c|}{\textbf{0.91 ($\pm$0.11)}} & \underline{\smash{0.79 ($\pm$0.10)}} \\ 
0.2        & \multicolumn{1}{c|}{0.75 ($\pm$0.08)} &  \multicolumn{1}{c|}{0.85 ($\pm$0.07)} & 0.72 ($\pm$0.08)\\ 
0.3      & \multicolumn{1}{c|}{0.78 ($\pm$0.09)} & \multicolumn{1}{c|}{0.82 ($\pm$0.31)} & 0.76 ($\pm$0.06)\\ 
0.5      & \multicolumn{1}{c|}{0.73 ($\pm$0.15)} & \multicolumn{1}{c|}{\underline{\smash{0.89 ($\pm$0.12)}}} & 0.76 ($\pm$0.07)\\ 
\end{tabular}
\end{table}

\newpage
\begin{figure}[!ht]
    \centering
    \includegraphics[width=12cm]{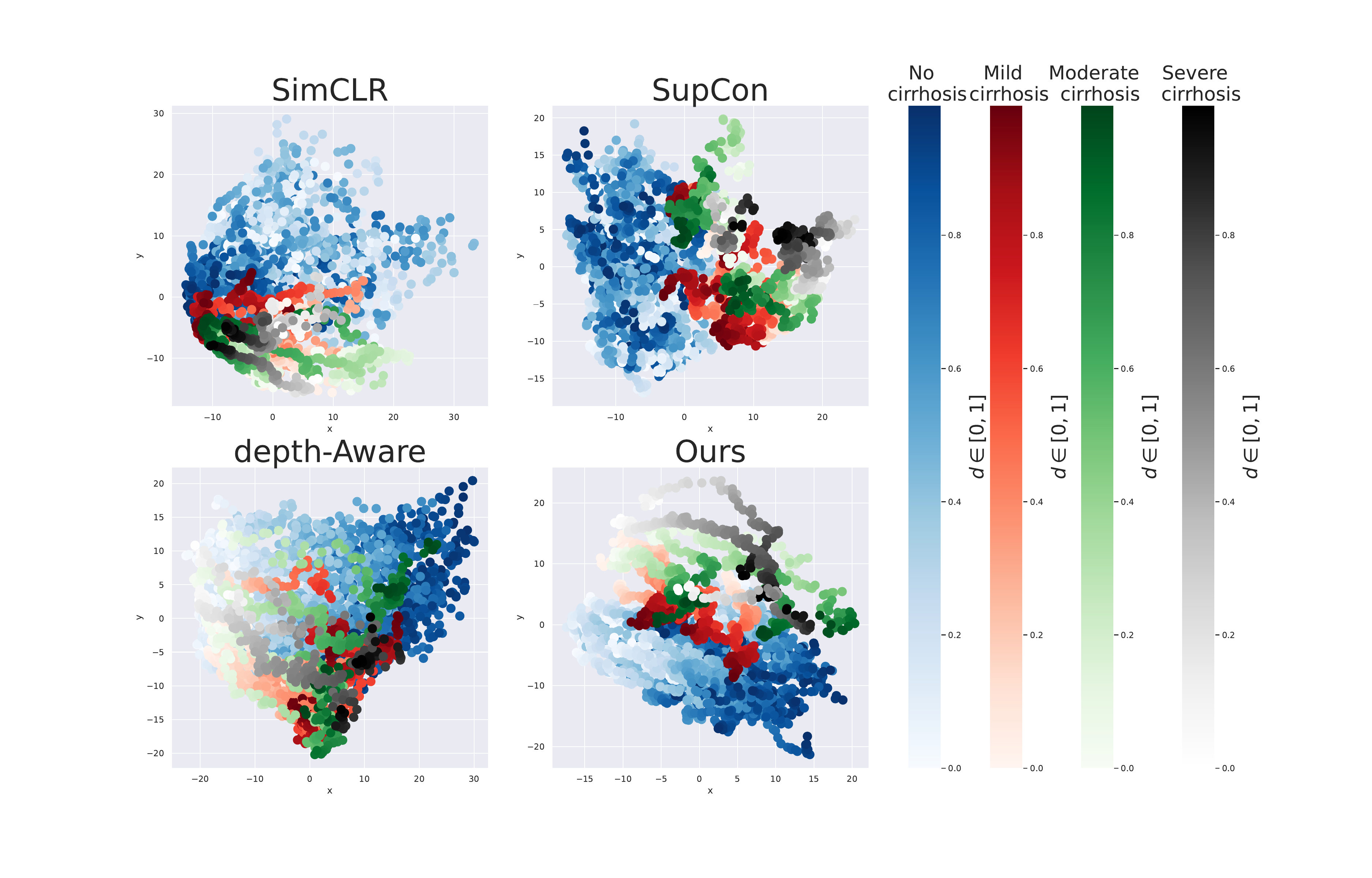}
    \caption{Projections of the ResNet-18 representation vectors of 70 randomly selected subjects of the training set of $\mathcal{D}_{radio}$ onto the first two modes of a PCA. Each dot represents a 2D slice. Color gradient refers to different depth positions. SimCLR and SupCon provide a remarkable separation between the healthy subjects (in blue) and the rest. However, classes mild moderate and severe are hardly separated. depth-Aware reaches an interesting global color gradient, but struggles to separate the cirrhotic classes. Our method provides the best class separation and at the same time correctly encodes the depth position.}
    \label{pca2}
\end{figure}

\begin{figure}[t]
     \centering
     \begin{subfigure}[t]{0.4\textwidth}
         \centering
         \includegraphics[width=4.5cm]{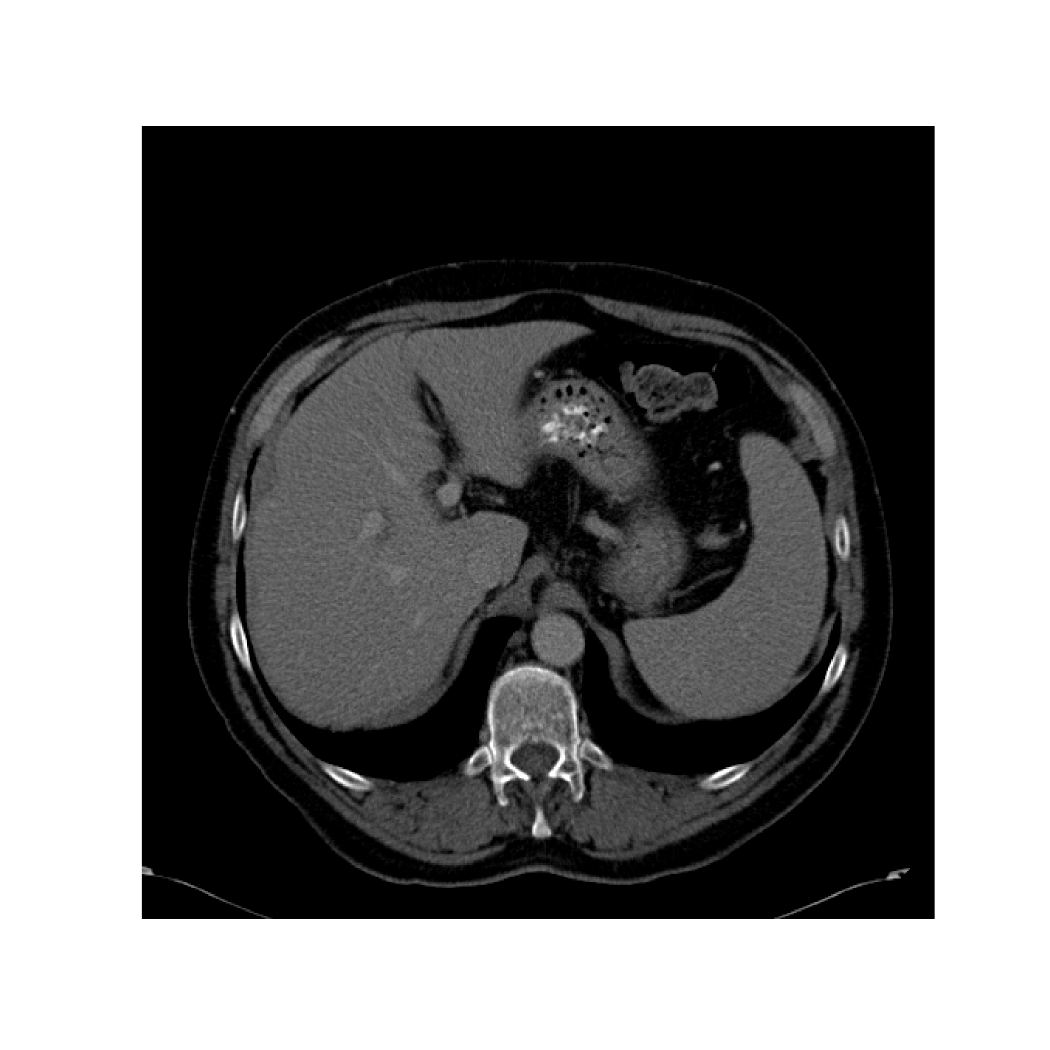}
         \caption{Cirrhotic case with the highest probability predicted by WSP.}
         \label{1}
     \end{subfigure}
     \begin{subfigure}[t]{0.4\textwidth}
         \centering
         \includegraphics[width=4.5cm]{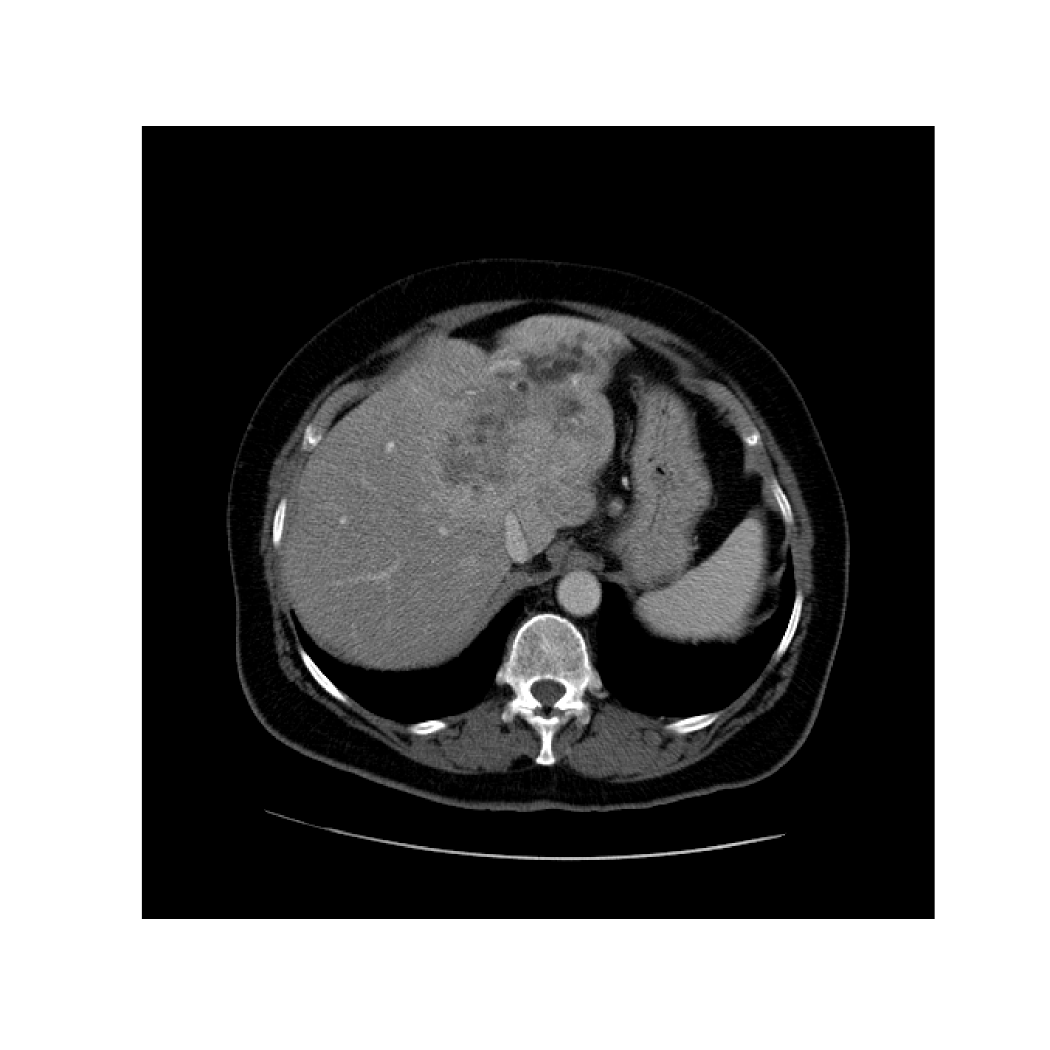}
         \caption{False negative misclassified by all the methods.}
         \label{2}
     \end{subfigure}
\caption{CT slices from $\mathcal{D}_{histo}^2$. On the left, the proposed method predicts the highest probability with 0.53 while SupCon, depth-Aware and SimCLR predict 0.51, 0.50 and 0.47 respectively. On the right, a false negative case predicted by all the models, possibly due to the slightly smaller size of the slice.}
\end{figure}

\end{document}